\documentclass{article}
\usepackage[utf8]{inputenc} 
\usepackage[margin=3.2cm]{geometry}

\usepackage{times}

\usepackage[varqu,varl,var0,scaled=0.97]{inconsolata}

\usepackage[protrusion=true,expansion=true]{microtype}

\usepackage{bm}
\usepackage{bbm}

\usepackage{parskip}

\usepackage[auth-lg,affil-sl]{authblk}

\author[1]{Jindong Gu}

\author[2]{Ahmad Beirami}

\author[2]{Xuezhi Wang}

\author[2]{Alex Beutel}

\author[1]{Philip Torr}

\author[2]{Yao Qin}

\affil[1]{University of Oxford}
\affil[2]{Google Research}

\date{}

\usepackage{graphicx}
\usepackage{amsmath}
\usepackage{amssymb}
\usepackage{amsfonts}
\usepackage{booktabs}

\usepackage{multirow}
\usepackage{caption}
\usepackage{subcaption}
\usepackage{array}
\usepackage{pbox}

\usepackage[square,sort,comma,numbers]{natbib}

\usepackage[T1]{fontenc}    

\usepackage{hyperref}       

\usepackage{url}            

\usepackage{booktabs}       

\usepackage{nicefrac}       

\usepackage{microtype}      

\usepackage{xcolor}         

\newcommand{\papertitle}{Towards Robust Prompts on Vision-Language Models}

\title{Towards Robust Prompts on Vision-Language Models}

\usepackage{fancyhdr}

\begin{document}

\maketitle

\begin{abstract}
With the advent of vision-language models (VLMs) that can perform in-context and prompt-based learning, how can we design prompting approaches that robustly generalize to distribution shift and can be used on novel classes outside the support set of the prompts? In this work, we first define two types of robustness to distribution shift on VLMs, namely, robustness on base classes (the classes included in the support set of prompts) and robustness on novel classes. Then, we study the robustness of existing in-context learning and prompt learning approaches, where we find that prompt learning performs robustly on test images from base classes, while it does not generalize well on images from novel classes. We propose robust prompt learning by integrating multiple-scale image features into the prompt, which improves both types of robustness. Comprehensive experiments are conducted to study the defined robustness on six benchmarks and show the effectiveness of our proposal.
\end{abstract}

\section{Introduction} 
\label{sec:intro}
Recently, large language models, such as GPT3~\cite{brown2020language}, have garnered great attention in the community due to their emergent abilities~\cite{wei2022emergent} that can be invoked via prompting. Due to the great success of prompt-based learning, recent works started to build large Vision-Language Models (VLMs) with such abilities for image understanding~\cite{tsimpoukelli2021multimodal, eichenberg2021magma, alayrac2022flamingo}. Specifically, visual features of images are first extracted and treated as visual tokens, and a frozen large LM outputs its understanding of the image by predicting the next text tokens conditioning on the known visual and textual tokens. 

Unlike that on language models~\cite{brown2020language,xie2021explanation,chan2022data,garg2022can, liu2021makes,lu2021fantastically,min2022rethinking}, \textbf{In-context Learning (IcoL)} on a frozen VLM, however, is highly sensitive to the demonstration examples~\cite{alayrac2022flamingo,eichenberg2021magma}. Thus, in order to achieve reasonable performance, the demonstration examples have to be elaborately selected from a support set.

Instead of selecting demonstration examples, an alternative method to utilize pre-trained VLMs is through \textbf{Prompt Learning (ProL)}~\cite{lu-etal-2022-fantastically,calibrate_before_use}, which employs labeled samples to generate embeddings that guide a language model in performing specific tasks. ProL replaces the demonstration examples' embeddings with learned embeddings, using the support set as a training dataset~\cite{brown2020language,zhou2022learning,zhou2022conditional}. The learned embeddings are then used for all query images during inferences.

Since both IcoL and ProL provide demonstration examples based on a small support set, there is a high chance that the query image in the inference comes from a shifted distribution. In addition, the test image can even belong to a novel class that does not exist in the support set. Under these different robustness scenarios for image classification, it is unclear which learning approach, in-context learning or prompt learning, is more robust against distribution shift.

To explore this question, in this work, we first define two types of robustness, namely, 1) robustness on base classes where the test images and the support set share the \textit{same} classes but come from different data distributions,  and 2) robustness on novel classes where the test images and the support have \textit{different} classes as well as different data distributions. To achieve different data distributions, we draw a list of images from ImageNet 1K~\cite{deng2009imagenet} as the support set, while the test images are obtained from the existing out-of-distribution robustness benchmarks~\cite{recht2019imagenet, hendrycks2021many, hendrycks2019robustness, wang2019learning, hendrycks2021nae}.

Then, we perform a study to compare the robustness of IcoL and ProL. Interestingly, we find that ProL performs more robustly on test images from base classes than IcoL, while it is worse at generalizing to images from novel classes. 
We explain this as: the retrieval-based in-context learning~\cite{yang2022empirical,alayrac2022flamingo}, where images most similar to the test image are selected from the support set as the demonstration examples, can hardly use all the available information in the support set. This results in poor generalization over distribution shifts. In contrast, prompt learning leverages all examples in the support set to learn more robust embeddings as demonstration examples, which significantly improves distributional robustness on \emph{base} classes. However, prompt learning runs a high risk of overfitting to the classes in the support set, leading to lower robustness to shifted images from \emph{novel} classes compared to in-context learning.

To address this, we further propose \textbf{robust ProL} by integrating multi-scale features of the image into the learned prompt, which is motivated by the strong robustness achieved by multi-scale network architecture designs~\cite{hendrycks2019robustness}. Specifically, multi-scale features of an image are taken as input in each forward pass when learning prompts for the language component of VLMs, which is feasible since they take a list of input tokens.
Extensive experiments show that our robust ProL can significantly improve robustness on both base and novel classes on six benchmarks.

Our contributions can be summarized as follows:
\begin{itemize}
    \setlength\itemsep{0.05em}
    \item We present the first comprehensive empirical study of the natural robustness issues of in-context learning (IcoL) and prompt learning (ProL) on visual-language models (VLMs) for image classification  on popular out-of-distribution (OOD) benchmark datasets.
    \item  Through extensive empirical evidence, we reveal that while existing ProL approaches offer better natural robustness on base classes that were present in the support dataset, IcoL generalizes better to novel unseen classes.
    \item Motivated by the robust multi-scale network architecture, we propose robust ProL that significantly improves robustness to both base and novel classes by integrating multi-scale features of an image into the prompt compared to existing IcoL and ProL approaches. We further provide an empirical analysis to better understand the learned robust prompt.
\end{itemize}

\section{Related Work}
\label{sec:related}

\noindent\textbf{Vision-language Models}.
Recent advance in vision and language modeling has been highly influenced by the advances in the language modeling domain~\cite{lu2019vilbert,li2019visualbert,su2019vl,chen2020uniter,li2020unicoder,kim2021vilt,jia2021scaling,alayrac2022flamingo,li2022dall}. Contrastive learning-based VLMs align visual features with text features extracted from LM, which show impressive zero-shot classification performance~\cite{radford2021learning,li2020unimo}. However, they can only be adapted to close-ended tasks, and it is also challenging to adapt contrastive models with a few examples~\cite{alayrac2022flamingo}. Instead of direct alignment, earlier VLMs based on BERT~\cite{devlin2019bert} extract object proposals from images with pre-trained object detectors and treat them as visual words, such as~\cite{tan2019lxmert,lu2019vilbert}. The BERT-based VLMs require fine-tuning when performing new tasks. Recent work builds VLMs on frozen GPT3-like LM where the in-context learning ability of LM can be retained. Specifically, Frozen~\cite{tsimpoukelli2021multimodal} trains a vision encoder to project input images to visual features that the language model can understand. Furthermore, MAGMA~\cite{eichenberg2021magma} further adds bottleneck adapters within the frozen LM to model visual tokens and text tokens in the same space. Instead of simple adapter layers, Flamingo~\cite{alayrac2022flamingo} integrates cross-attention layers interleaved between the frozen LM layers to further improve model performance. The VLMs based on GPT3-like LM can perform new tasks without finetuning the model parameters. In this work, we focus on these Flamingo-like VLMs.

\noindent\textbf{In-context Learning}.
In-context learning describes a model's ability to perform a given task just by conditioning on demonstration examples, e.g., a few input-output examples~\cite{xie2021explanation}. The concept of IcoL was popularized by GPT-3 where the model completes a task instance by predicting what comes next, conditioning on a natural language instruction, and/or a few demonstrations of the task~\cite{brown2020language}. More interestingly, large-scale LM based on this concept can achieve quite impressive performances based on just a few examples on various NLP benchmarks~\cite{brown2020language}. Recently, in-context learning ability has also been explored to understand images with the assistance of natural languages. Specifically, existing works build vision-language models on top of a strong frozen language model to retain the ability of in-context learning~\cite{tsimpoukelli2021multimodal,eichenberg2021magma,yang2022empirical,alayrac2022flamingo}.

\noindent\textbf{Prompt Learning}.
Despite the rapid progress of in-context learning in the language domain, recent work showed that language models' performance can be highly sensitive to the input prompts~\cite{lu-etal-2022-fantastically,calibrate_before_use}, thus proposing methods to learn better prompts such as prompt tuning~\cite{prompt_tuning} and automatic prompt search~\cite{better_few_shot,autoprompt:emnlp20,rubin-etal-2022-learning}. Similarly for VLMs, during inference time for contrastive-learning-based VLMs~\cite{radford2021learning}, current literature also proposes to learn the \textit{virtual} text as part of texts. For example, CoOp learns part of the prompt with input-output pairs~\cite{zhou2022learning}. CoCoOp improves the performance on novel classes by conditioning the prompt learning on visual features with a meta-network~\cite{zhou2022conditional}. Besides, the model performance can be also improved with an Adapter in visual or textual encoder~\cite{zhang2021tip,gao2021clip}. However, both CoCoop and Adapter often require many input-output pairs to finetune a Meta-Net or an Adapter layer. Currently, prompt learning has not been explored on Flamingo-like VLMs.

\noindent\textbf{Natural Robustness}.
In real-world applications, the text samples can be domain-shifted from the training data distribution. Various robustness benchmark datasets have been proposed to test model performance on distributional-shifted images, i.e., ImageNet-V2~\cite{recht2019imagenet}, ImageNet-R~\cite{hendrycks2021many}, ImageNet-C~\cite{hendrycks2019robustness}, ImageNet-S~\cite{wang2019learning} and ImageNet-A~\cite{hendrycks2021nae}. Robustness on the benchmarks~\cite{taori2020measuring,hendrycks2019robustness,gu2022vision,wu2022towards} and various network architectures~\cite{hendrycks2019robustness,gu2020improving,gu2021capsule,gu2022vision} has been intensively studied in deep classifiers trained in supervised learning. Recently, the robustness of various foundation models raises the attention of community.~\cite{bommasani2021opportunities,tran2022plex,nguyen2022quality,fan2021does,chen2020adversarial}. Given the importance of in-context learning~\cite{brown2020language,alayrac2022flamingo}, in this work, we make the first exploration of the robustness of IcoL and ProL on Flamingo-like VLMs.

\section{Formulating Robustness of Image Classification on VLMs}
In this section, we first present Image Classification on a Flamingo-like VLM and define two types of robustness.

\noindent\textbf{Image Classification on VLMs with frozen LM and VE}. Flamingo, and several other similar VLMs, are built on a frozen large LM and a Visual Encoder (VE). The features extracted from VE are formulated into visual tokens. Together with text tokens, visual tokens are fed to a LM to obtain LM's outputs.

Formally, an input of the VLM consists of context information $\bm{C}$, a test image $\bm{X}^t$, and query texts $\bm{T}^q$, which can be described as:
\begin{equation}
\textit{Input} = (\bm{C}, \; \bm{X}^t, \; \bm{T}^q ) \\
\label{euq:input}
\end{equation}
where the context information $\bm{C}$ can be a few image-text pairs $\{\bm{X}_k, \bm{T}_k\}$ sampled from a support set (in IcoL) or learned virtual tokens (in ProL), $\bm{X}^t$ is the test image the VLM is expected to describe, and $\bm{T}^q$ is the query text that gives a hint to LM.

The input above is first processed and converted into a list of visual and text tokens:
\begin{equation}
\textit{Tokens} = (\bm{C}', \; \bm{V}^t, \; \bm{W}^q) \\
\label{euq:token}
\end{equation}
where $\bm{V}^t$ the visual embedding (e.g. a list of embedding tokens) of the test image $\bm{X}^t$ extracted from VE, and $\bm{W}^q = \{\bm{W}^q_{j}\}$ is a list of textual tokens of texts $\bm{T}^q$. $\bm{C}'$ are learnable tokens or tokens extracted from image-text pairs.

The LM component of the VLM predicts the next token conditioning on the previous tokens:
\begin{equation}
\textit{Pred} = P(W^q_{j+i} \; \vert \; \bm{C}', \; \bm{V}^t, \; \bm{W}^q) \\
\end{equation}

\begin{figure*}[t]
\centering
\includegraphics[scale=0.18]{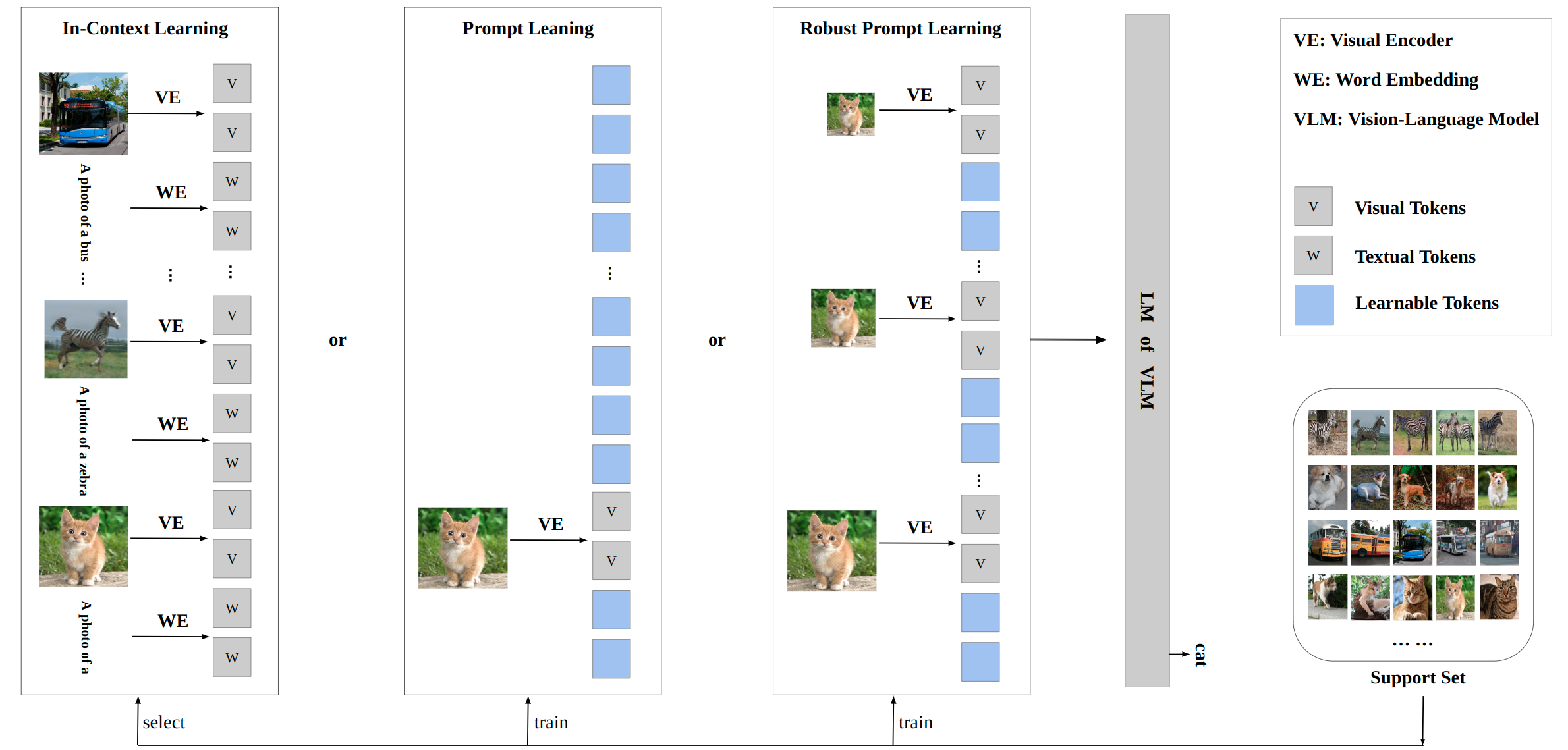}
\caption{Overview of different learning approaches on VLMs. 1) IcoL: Contextual examples are selected manually from a support set, e.g., randomly~\cite{eichenberg2021magma} or by visual feature-based retrieval~\cite{yang2022empirical,alayrac2022flamingo}. Query texts are also manually designed, e.g., ``A photo of a". 2) ProL: We leverage all the examples in the support set to learn a virtual prompt. 3) Robust-ProL: We integrate multi-scale visual features into the learnable prompt to make it robust. Conditioned on the input prompt, the LM in the VLM predicts the class corresponding to the last image. VE, WE, V and W mean Visual Encoder, Word Embedding, Visual Tokens, and Textural Tokens, respectively.} 
\label{fig:overview}
\end{figure*}

Given a set of labels $\mathcal{Y}$, a text image can be classified with the following equation:
\begin{equation}
\bm{Y}^{t,p}(\bm{X}^t) = \underset{\bm{Y}_i \in \, \mathcal{Y}}{\mathrm{argmax}} \;  P( \bm{Y}_i \; \vert \; \bm{C}', \; \bm{V}^t, \; \bm{W}^q) \\
\end{equation}
Following~\cite{eichenberg2021magma,alayrac2022flamingo}, we append all possible labels independently to the query text, and each of the resulting sequences will be scored using the log-likelihood estimated by the LM. The one with the highest confidence is the predicted class. For example, as shown in Fig.~\ref{fig:overview}, the \textit{cat} is most likely to be the next predicted token.

The context information $\bm{C}$ can be a list of image-text pairs sampled from a support set or learnable tokens optimized on the support set. In practice, the test image $\bm{X}^t$ is likely to deviate from the sample distribution of the support set since only a few examples are available therein. Besides, the label set of test images can even be different from the ones in the support set. Given the observation, we define the following two types of robustness.

\noindent\textbf{Robustness on Base Classes}. The support set includes a few image-text pairs (i.e., labeled images) per class. All classes included in the support set are defined as base classes $\mathcal{Y}$. Robustness on Base Classes is defined as the model robustness to the distribution shift of the test images from base classes. 

The robustness on base classes is defined as $\mathbb{E}_{Y^t\in \mathcal{Y}} [\mathbbm{1}(\bm{Y}^{t,p}(\bm{X'}^t) = \bm{Y}^t)]$ where the ground-truth label is associated with the distribution-shifted image, $\bm{X'}^t,$ is $\bm{Y}^t$. All test images share the same label set $\mathcal{Y}^t = \mathcal{Y}$.

\noindent\textbf{Robustness on Novel Classes}. The test images on VLM can be drawn from novel classes that are not included in the support set. We define robustness on novel classes as the model robustness to the distribution shift of the test images from novel classes. In this case, the label set $\mathcal{Y}^t$ of test images has no overlap with the base class set $\mathcal{Y}$.
$\mathbb{E}_{Y^t\in \mathcal{Y'}} [\mathbbm{1}(\bm{Y}^{t,p}(\bm{X'}^t) = \bm{Y}^t)]$
where the test images and the samples of the support set are from the disjoint label sets, i.e., $\mathcal{Y}^t \cap \mathcal{Y} = \emptyset$.

\section{Methodology}
\subsection{In-Context Learning on VLMs}
VLMs with IcoL ability are able to recognize a test image just by conditioning on a few input-output examples. In other words, the context information in Equation~\ref{euq:input} is $\bm{C} = (\bm{X}_1, \bm{T}_1, ..., \bm{X}_K, \bm{T}_K)$ where the $K$ demonstration examples are sampled from a support set. Before being fed to the LM component of the VLM, the selected examples are first processed into tokens $\bm{C}' = (\bm{V}_1, \bm{W}_1, ..., \bm{V}_K, \bm{W}_K)$ where $\bm{V}_k = \{\bm{V}_{kj}\}$ is the visual embedding (e.g. a list of embedding tokens) of the image $\bm{X}_k$ extracted from VE, and $\bm{W}_k = \{\bm{W}_{kj}\}$ is a list of textual tokens of texts $\bm{T}_k$.

As illustrated in the first column (In-context Learning) of Fig.~\ref{fig:overview}, to perform image classification, the VLM predicts the text token ``\textit{Cat}" conditioned on a few contextual examples (image-text pairs where the text describes the image class), a test image of \textit{Cat} and a query text ``\textit{A photo of a}".

The IcoL on VLM is highly sensitive to the selected demonstration samples. Previous work selects examples either randomly~\cite{eichenberg2021magma} or b Retrieval-based In-Context Example Selection (RICE) from a support set~\cite{yang2022empirical,alayrac2022flamingo}. While random selection does not work well, IcoL-RICE achieves reasonable performance by comparing the visual features of images from the support set with the query image features and choosing the most similar ones as demonstrations.

\subsection{Prompt Learning on VLMs}
Compared to previous methods of manually selecting contextual examples as well as query texts, ProL, instead, makes both of them become learnable embeddings. To optimize these learnable embeddings, all the examples from the support set are applied to maximize the likelihood of the following sequence on the LM
\begin{equation}
P( \bm{L}^c, \; \bm{V}_i, \; \bm{L}^q ,\;  \bm{Y}_i) \\
\end{equation}
where $\bm{V}_i$ is the visual embedding of the image $\bm{X}_i$ from the support set and $\bm{Y}_i$ is the corresponding image class. $\bm{L}^c$ and $\bm{L}^q$ represent the learnable contextual embedding and query text embedding respectively, whose length is set as a hyperparameter. All learnable embeddings are randomly initialized with $\mathcal{N}(0, 1)$, in order that the learned prompt does not biased to any manual design. The learnable embeddings are optimized through a standard gradient-based optimizer with regard to the above training objective. During inference, the $\bm{V}_i$ will be replaced by the embedding of the test image $\bm{V}^t$, and the model is expected to output the corresponding image class $\bm{Y}^t$ conditioning on the learnable embeddings and visual feature of the text image, i.e., $(\bm{L}^c,\, \bm{V}^t, \, \bm{L}^q)$. ProL improves the robustness on base classes, while it does not work well on the robustness on novel classes, as we discussed later in Sec.~\ref{sec:exp_novel}.

\begin{figure*}[!ht]
    \centering
    \includegraphics[scale=0.6]{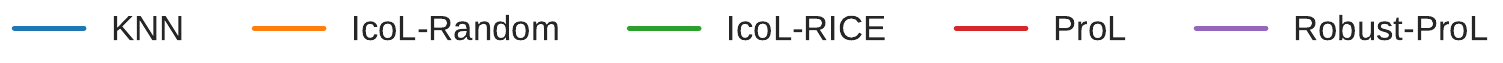}

    \begin{subfigure}[b]{0.3\textwidth}
        \includegraphics[scale=0.43]{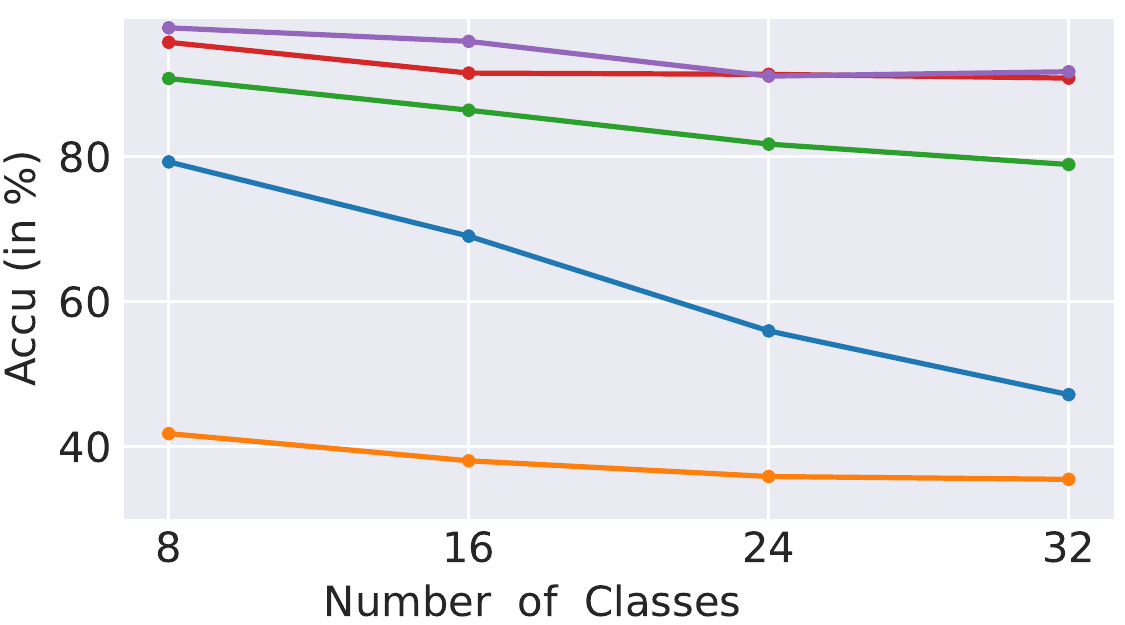}
    \caption{ImageNet}
    \label{subf:imagenet}
    \end{subfigure} \hspace{5mm}
    \begin{subfigure}[b]{0.3\textwidth}
        \includegraphics[scale=0.43]{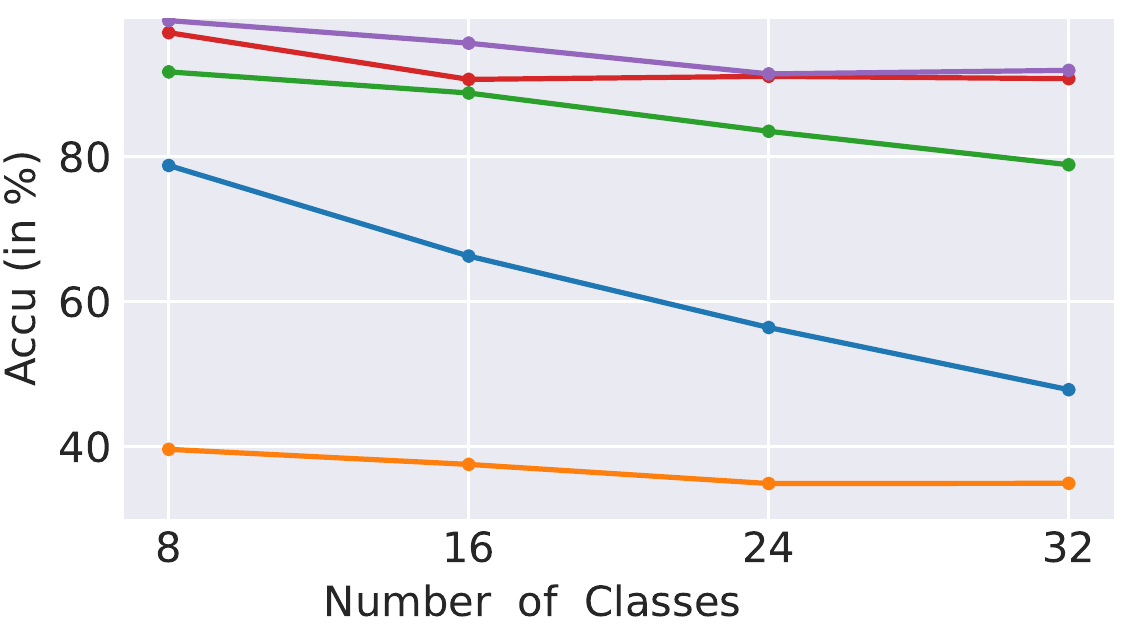}
        \caption{ImageNet-V2}
        \label{subf:imagenet_v2}
    \end{subfigure} \hspace{5mm}
    \begin{subfigure}[b]{0.3\textwidth}
        \includegraphics[scale=0.43]{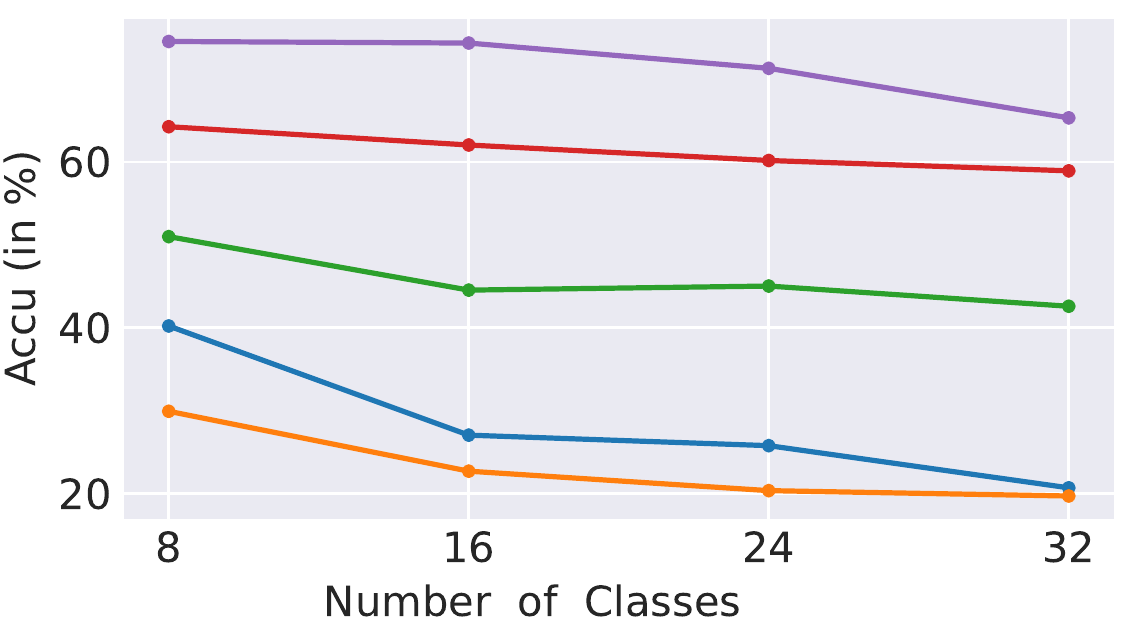}
        \caption{ImageNet-R}
        \label{subf:imagenet_r}
    \end{subfigure}

    \begin{subfigure}[b]{0.3\textwidth}
        \includegraphics[scale=0.43]{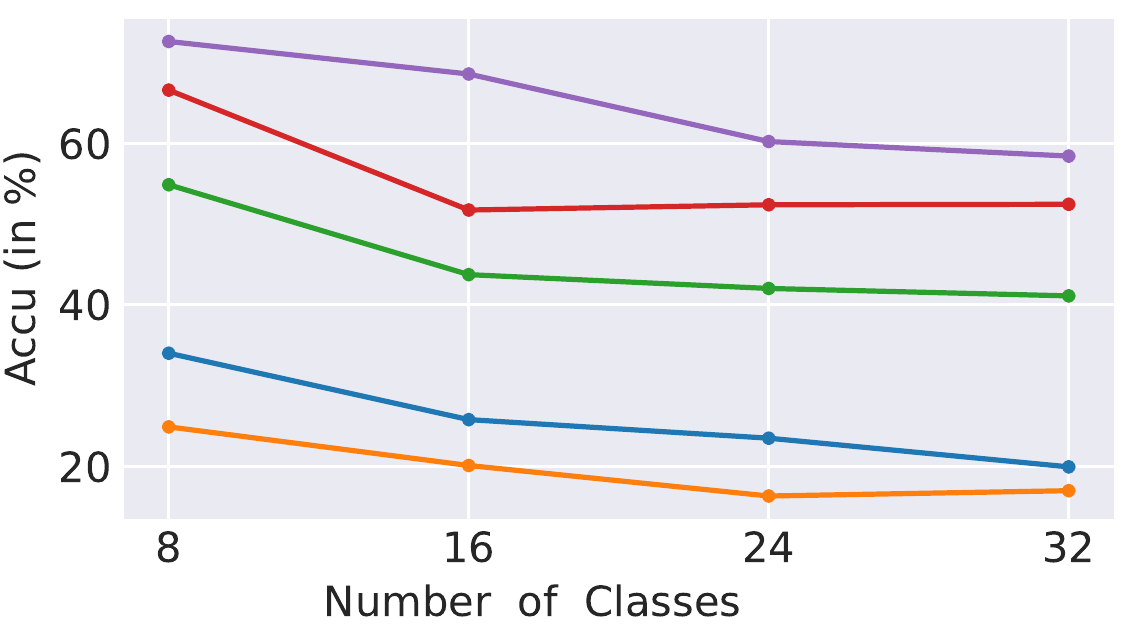}
        \caption{ImageNet-C}
        \label{subf:imagenet_c}
    \end{subfigure} \hspace{5mm}
    \begin{subfigure}[b]{0.3\textwidth}
        \includegraphics[scale=0.43]{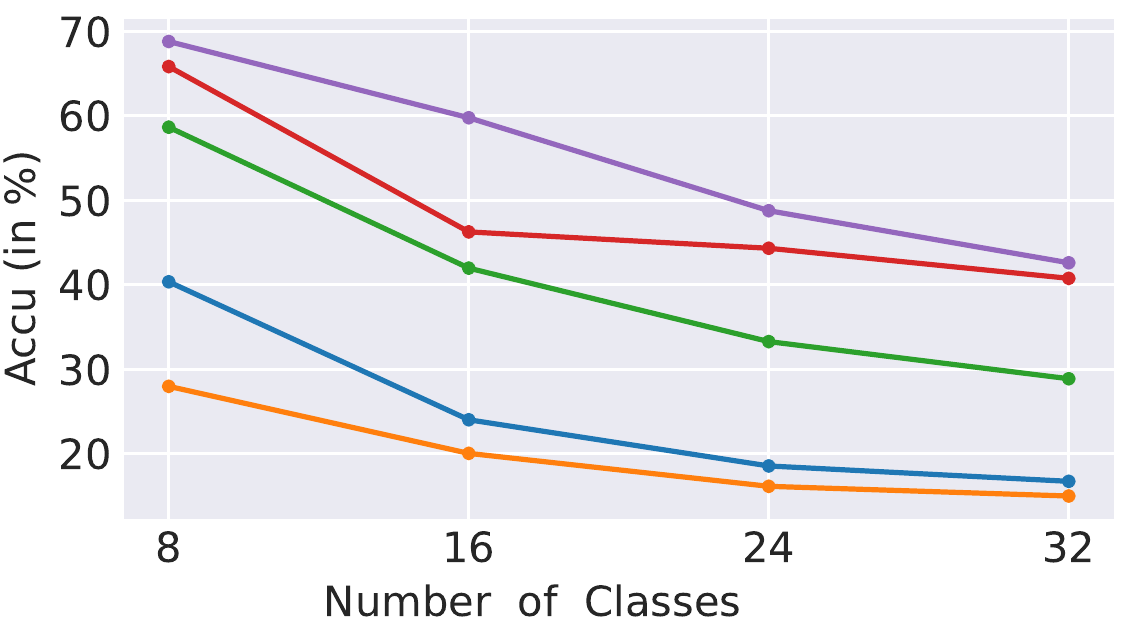}
        \caption{ImageNet-S}
        \label{subf:imagenet_s}
    \end{subfigure} \hspace{5mm}
    \begin{subfigure}[b]{0.3\textwidth}
        \includegraphics[scale=0.43]{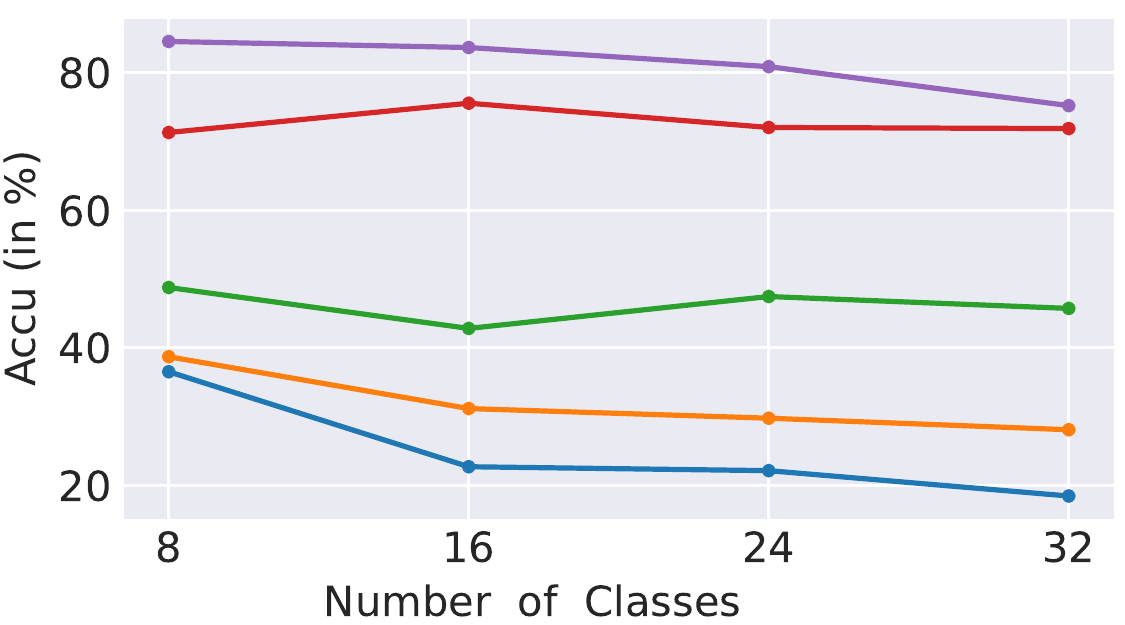}
        \caption{ImageNet-A}
        \label{subf:imagenet_a}
    \end{subfigure} 
    \caption{VLM performance on base classes on six datasets. The base classes are the classes used in the support set. 
    The first subplot shows the in-distribution performance (the support set and the test set are both from ImageNet), while the rest corresponds to the out-of-distribution performance (the test set comes from each OOD dataset). In each subplot, the model classification accuracy of different methods is shown in different settings, i.e., classification of 8, 16, 24, and 32 classes. As shown, Robust-ProL achieves similar performance to ProL on the in-distribution ImaegNet dataset, both of which outperform others. On OOD datasets ImageNet-R/C/S/A, Robust-ProL outperforms all other methods significantly.}
    \label{fig:robust_base} 
\end{figure*}

\subsection{Robust Prompt Learning on VLMs}
In this section, we present our proposed method for designing robust prompt learning.
Previous work finds that multi-scale network architecture designs improve the recognition robustness by extracting multi-scale image features in forward pass~\cite{hendrycks2019robustness}. Motivated by the findings, we propose to integrate multiple-scale image features into the input, which is feasible given the VLM's ability that many images can be considered together in a single forward pass. In this way, VLMs can leverage image features of multiple scales to make a prediction, which makes the predictions robust to distribution shifts. 
Concretely, the robust prompt with multi-scale features can be described as 
\begin{equation}
P(\bm{V}^{s1}_i,\; \bm{L}^{c1},\; \bm{V}^{s2}_i,\; \bm{L}^{c2},\; \bm{V}^{s3}_i, \; \bm{L}^q ,\; \bm{Y}_i) \\
\end{equation}
where $\bm{V}^{s1}_i, \bm{V}^{s2}_i, \bm{V}^{s3}_i$ are visual features of the same image $\bm{X}_i$ at different scales. 
Three popular sizes (224, 299, 384) are considered in this work where 384 is the one used in pre-training. The learnable embeddings $\bm{L}^{c}$ in ProL is evenly split into 2 pieces $\bm{L}^{c1}$ and $\bm{L}^{c2}$ to be paired with the visual feature embeddings $\bm{V}^{s1}_i$ and $\bm{V}^{s2}_i$. We optimize the learnable embeddings $\bm{L}^{c1}$, $\bm{L}^{c2}$ and $\bm{L}^q$ following the same way as ProL.
Note that the length of learnable embeddings for ProL and Robust-ProL is set to be the same. As shown in Sec.~\ref{sec:exp_novel}, Robust-ProL significantly improves the robustness of VLM on both base classes and novel ones.

\section{Experiments}
\label{sec:exp}

\noindent\textbf{Model.} The VLM we use is MEGMA~\cite{eichenberg2021magma}, which is a publicly available vision-language model with in-context learning ability. MEGMA is composed of a visual encoder (VE) NF-RN50x16~\cite{brock2021high} of and a frozen language model (LM) GPT-Neo~\cite{gpt_neo} with adapters of 2.7B. In inference, the visual features extracted with the VE are treated as visual words. The image-caption pairs are first processed as a list of tokens and then fed to the adapted LM. The model predicts the next token in an auto-regressive manner.

\noindent\textbf{Datasets.} Six datasets are used in our experiments. ImageNet-1k~\cite{deng2009imagenet} is taken as our in-distribution data, and ImageNet-V2~\cite{recht2019imagenet} (re-collected ImageNet-like images), ImageNet-R~\cite{hendrycks2021many} (rendition images), ImageNet-C~\cite{hendrycks2019robustness} (natural corrupted images), ImageNet-S~\cite{wang2019learning} (sketch images), ImageNet-A~\cite{hendrycks2021nae} (natural adversarial images) are tested as OOD data. We use subsets of these datasets by selecting images from the classes shared by all six datasets (86 classes are shared). The names of both base and novel classes are in Appendix A.

\noindent\textbf{Tasks.} We focus on classification tasks since \textit{popular robustness benchmark datasets are ImageNet-variant}. Given that there are 86 joint classes, we consider the following settings, i.e., classification of 8, 16, 24, and 32 classes respectively. 
We randomly sample 8, 16, 24, or 32 classes from the joint classes and use them for all the methods across all datasets for fair comparisons. Following~\cite{alayrac2022flamingo}, we use 5 labeled images per class for in-context learning. 

\begin{table}[t] 
\begin{center}
\small
\setlength\tabcolsep{0.06cm}
\begin{tabular}{c|  cccc | c}
\toprule
Methods  & Prompt & VE & Linear-Cls & LM  & \#Samples \\
\midrule
Supervised$^*$  & - & Frozen  & Learnable & -  &  all*N  \\
\midrule
\midrule
Supervised-FS  & - & Frozen  & Learnable  & -  & 5N  \\
\midrule
IcoL-Random & Random &  Frozen & - & Frozen  &  5N  \\
\midrule
IcoL-RICE & Retrieval &  Frozen & - & Frozen  &  5N  \\
\midrule
ProL  & Learnable &  Frozen & -  & Frozen  &  5N  \\
\midrule
Robust-ProL  & Learnable &  Frozen & -  &  Frozen &  5N \\
\bottomrule
\end{tabular} \vspace{-0.15cm}
\end{center}
\caption{We introduce two supervised baselines: 1) Supervised$^*$: All images of target classes in ImageNet are applied to fine-tune a learnable linear layer on a frozen VE of the same VLM. 2) Supervised-FS: only a few examples per class are available for linear finetuning.} 
\label{tab:supervised}
\end{table}

\begin{table*}[t]
\begin{center}
\footnotesize
\setlength\tabcolsep{0.06cm}
\begin{tabular}{c| ccc| ccc| ccc| ccc| ccc| ccc}
\toprule
 Datasets & \multicolumn{3}{c|}{ImageNet} & \multicolumn{3}{c|}{ImageNet-V2} & \multicolumn{3}{c|}{ImageNet-R} &  \multicolumn{3}{c|}{ImageNet-C} &  \multicolumn{3}{c|}{ImageNet-S} & \multicolumn{3}{c}{ImageNet-A} \\
\midrule
No. of Classes & 8 & 16 & 32  &  8 & 16 & 32  & 8 & 16 & 32 & 8 & 16 & 32 & 8 & 16 & 32 & 8 & 16 & 32 \\
 \midrule
Supervised$^*$   & \textbf{99.50} & \textbf{98.37} & \textbf{96.06}   & \underline{97.92} & \textbf{97.08} & \textbf{95.31}   & \textbf{77.43} & \underline{71.59} & \underline{64.14}   & \textbf{74.12} & \textbf{71.63} &  \textbf{63.23}  & \underline{83.58} & \underline{83.07} & \textbf{77.32}   & \underline{67.08} & \underline{57.07} & \underline{41.64} \\
\midrule
\midrule
Supervised-FS  & 96.75 & 92.12 &  85.87  & 94.17 & 91.88 & 86.56  & 66.14 & 46.59 & 43.88 & 64.53 & 59.14 & 39.69 & 64.22 & 42.33 & 40.38 & 61.88 & 38.63 & 30.01
\\
\midrule
KNN   & 79.25 & 69 & 47.13  & 78.75 & 66.25 & 47.81  & 40.2 & 27.02 & 20.66   & 34.01 & 25.79 & 19.94  & 36.52 & 22.7 & 18.44   & 40.35 & 24.01 & 16.72  \\
\midrule
IcoL-Random  & 41.75 & 38 & 35.44   & 39.58 & 37.5 & 34.89   & 29.9 & 22.67 & 19.67  & 24.89 & 20.11 & 16.99  & 38.7 & 31.16 & 28.08  & 27.97 & 20.03 & 14.98 \\
\midrule
IcoL-RICE      & 90.75 & 86.38 & 78.88 & 91.67 & 88.75 & 78.85 & 50.99 & 44.53 & 42.58 & 54.87 & 43.73 & 41.11 & 48.77 & 42.82 & 45.73 & 58.66 & 41.97 & 28.87 \\
\midrule
ProL  & 95.75 & 91.5  & 90.81 & 97.08  & 90.63 & 90.73 & 64.26 & 62.05  & 58.93  & 66.59 & 51.74 & 52.45 & 71.32 & 75.58 & 71.89 & 65.84 & 46.26 & 40.76 \\
\midrule
Robust-ProL  &  \underline{97.75} & \underline{95.88} & \underline{91.69} & \textbf{98.75}  & \underline{95.63} & \underline{91.88} & \underline{74.55} & \textbf{74.36} &  \textbf{65.33} & \underline{72.61} & \underline{68.58} & \underline{58.42}  & \textbf{84.56} & \textbf{83.68} & \underline{75.23} & \textbf{68.81} & \textbf{59.78} & \textbf{42.6}\\
\bottomrule
\end{tabular} \vspace{-0.15cm}
\end{center}
\caption{Robustness on base classes: The bold means the best, and the underline corresponds to the second best. Robust ProL outperforms the Supervised-FS baseline on both in-distribution and OOD datasets. Compared with the Supervised$^*$ baseline (trained on 10306 images), Robust ProL trained on 5N images, where N is the number of classes, achieves slightly lower accuracy on in-distribution ImageNet, while obtaining comparable robustness on OOD datasets.}
\label{tab:supervised_results}
\end{table*}

\noindent\textbf{Baselines.}
1) \textit{KNN}: K-Nearest Neighbor in visual feature space is a baseline since in-context learning assumes a few labeled images available. K is set to 8 for comparison.
2) \textit{IcoL-Random}: Following~\cite{eichenberg2021magma}, we randomly sample 8 labeled images (i.e. image-class name pairs) and take them as demonstrations.
3) \textit{IcoL-RICE}: Following~\cite{yang2022empirical,alayrac2022flamingo}, we select 8 labelled images by visual feature-based retrieval. Given a test image, we compare its visual feature extracted by VE with the ones of all available labeled images and choose the most similar 8 images as demonstrations.
4) \textit{ProL}: We set the number of the demonstration examples token embeddings to 1024, which is similar to the one in manual demonstrations (8 images + 8 descriptions). 
5) \textit{Robust-ProL}: We integrate three scale features of the test image into the context, namely, 224, 299, and 384. We set 512 learnable tokens respectively after the first two visual tokens. The last scale (384) is the same as used in all previous methods.
For both ProL methods, the number of learnable tokens of the query text is set to 64, and the tokens are randomly initialized with normal distribution $\mathcal{N}(0, 1)$. The context is tuned with a batch size of 16 for 50 epochs using SGD with a learning rate of 1.0. The ablation studies on the hyper-parameters are in Appendix B.

\subsection{Robustness on Base Classes}
\label{sec:basec}
In this subsection, we study model performance when test images share the class label set with the support set.
The performance is reported in Fig.~\ref{fig:robust_base}. Each of the six subplots corresponds to a dataset. Fig.~\ref{subf:imagenet} on ImageNet shows the in-distribution performance where both the support set and test images are from the same data distribution (i.e., the same dataset). In Fig.~\ref{subf:imagenet_r}-\ref{subf:imagenet_a}, we report model performance on test images from various OOD datasets where we still take the same support set from ImageNet. Namely, the learnable prompt is trained on the images from ImageNet dataset and tested on the images from OOD datasets.

In each subplot, the x-axis shows the number of classes, we consider four settings, i.e., classification of 8, 16, 24, and 32 classes. The classes and the corresponding support set are pre-selected randomly and fixed across all the experiments. The y-axis shows the model performance, i.e., classification accuracy in percentage. Each line corresponds to a classification method.

In Fig.~\ref{subf:imagenet}, we can observe that ProL and Robust-ProL show similar performance, and both two outperform others significantly. The learned virtual embedding outperforms the retrieval-based in-context example selection (IcoL-RICE), while the IcoL-RICE outperforms random selection significantly. In Fig.~\ref{subf:imagenet_v2}, the test images from ImageNet-v2 are collected in the same way as ImageNet. Previous work~\cite{recht2019imagenet} finds that the performance on ImageNet-v2 is lower than that on ImageNet. Different from previous work, we do not observe any performance degradation in our learning setting. In Figs.~\ref{subf:imagenet_r}-\ref{subf:imagenet_a}, ProL shows clearly better performance than IcoL-RICE and random baselines on the images from ImageNet-R/C/S/A, and Robust-ProL outperforms ProL significantly in various settings. Overall, ProL shows higher robustness on base classes, and Robust-ProL improves the robustness further.

\begin{table*}[t]
\begin{center}
\footnotesize
\setlength\tabcolsep{0.03cm}
\begin{tabular}{c| cc| cc| cc| cc| cc| cc}
\toprule
  & \multicolumn{2}{c|}{ImageNet} & \multicolumn{2}{c|}{ImageNet-V2} & \multicolumn{2}{c|}{ImageNet-R} &  \multicolumn{2}{c|}{ImageNet-C} &  \multicolumn{2}{c|}{ImageNet-S} & \multicolumn{2}{c}{ImageNet-A} \\
\midrule
 & Base Cls  & Novel Cls  &  Base Cls  & Novel Cls  & Base Cls  & Novel Cls & Base Cls  & Novel Cls & Base Cls  & Novel Cls & Base Cls & Novel Cls \\
 \midrule
Unsupervised  & 80.75 & \textbf{82.7} & 82.92 & \textbf{82.55} & \textbf{67.52} & \textbf{64.46}  & 62.38 & \textbf{69.25} & \underline{72.30} &  \textbf{76.29} & 62.87 & \underline{53.12} \\
\midrule
\midrule
IcoL-Random  & 41.75 & 68.5  & 39.58 & 70.92  & 29.9 & 47.47 & 24.89 & 40.46  & 38.7  & 64.19 & 27.97 & 36.77 \\
\midrule
IcoL-RICE  & 90.75  &  66 & 91.67 & 69.58 & 50.99 & 45.59 & 50.99 & 40.92  & 48.77  &  61.23 & 58.66  & 41.29 \\
\midrule
ProL  & \underline{95.75} & 43.5  & \underline{97.08} & 45.42  & 64.26 & 39.56 & \underline{66.59} & 39.23  & 71.32  & 40.99 & \underline{65.84}  & 20.65 \\
\midrule
Robust-ProL  &  \textbf{97.75} & \underline{73.25} & \textbf{98.75} & \underline{71.25}  & \textbf{74.55} & \underline{64.27} & \textbf{72.61} & \underline{60.28}   &  \textbf{84.56} & \underline{64.94} & \textbf{68.81} & \textbf{53.76}  \\
\bottomrule
\end{tabular} 
\end{center}
\caption{Robustness on novel classes: The bold marks the best, and the underline marks the second best. Performance on base classes is included for comparison. Robust-ProL outperforms all other methods on both base classes and novel ones.}   
\label{tab:novel}
\end{table*}

\noindent\textbf{Comparison to Robustness of Supervised Learning.} We also compare the robustness IcoL and ProL on VLM to standard supervised learning. Since only a few samples are available in the support set, it is infeasible to train a large network from scratch based on the same support set. Hence, we 
finetune a learnable linear classifier on top of the frozen VE from the same VLM with 5N examples, where N is the number of classes. We denote this model as Supervised-FS (few-shot). Note that the same set of examples is used to learn the linear classifier in Supervised-FS (few-shot) as well as in ProL and Robust-ProL for a fair comparison. Furthermore, we also report the performance of a stronger supervised baseline, denoted as ``Supervised$^*$'', where the linear classifier is finetuned using all images belonging to the N classes in ImageNet, which in total is 10306 images. All the models in Tab.~\ref{tab:supervised} are built on the same frozen VE for a fair comparison, although Supervised$^*$ takes in much more examples compared to others (10306 vs. 5N images).

As shown in Tab.~\ref{tab:supervised_results}, ProL performs similarly as Supervised-FS on a small number of classes and outperforms Supervised-FS when more classes are considered. This is mainly because Supervised-FS is fine-tuned with a few labeled examples, which is not sufficient to support highly accurate image classification on more classes. In addition, Robust-ProL outperforms both ProL and Supervised-FS by a large margin on OOD datasets ImageNet-R/C/S/A. Lastly, Supervised$^*$ baseline outperforms our Robust-ProL on the in-distribution dataset. However, Robust-ProL can outperform Supervised$^*$ on many OOD datasets (ImageNet-R/S/A). Note that Supervised$^*$ is not a fair baseline since it takes in much more training data compared to Robust-ProL (10306 vs. 5N images). This strongly supports the effectiveness of our Robust-ProL in improving models' robustness on OOD datasets.

\subsection{Robustness on Novel Classes}
\label{sec:exp_novel}
In this subsection, we study model performance on test images from novel classes.
The performance is reported in percentage in Tab.~\ref{tab:novel}. For each method on each dataset, we report two scores, i.e., the accuracy on base classes and the one on novel classes. The accuracy on base classes is shown in the table for comparison where the test images and the support set share the same class set. In contrast, the accuracy on novel classes means that the classes of test images are different from the ones in the support set.

On ImageNet dataset, both the test images and the support set are taken from ImageNet dataset although they are from different classes. For other datasets, we use the same support set, but the test images of novel classes are taken from OOD datasets. Note that the same novel classes are used across all datasets.

As shown in Tab.~\ref{tab:novel}, ProL significantly improves the robustness on base classes. but its accuracy degrades significantly when the test images are coming from novel classes, even worse than IcoL-Random and IcoL-RICE baselines. This suggests that simply applying ProL is insufficient for strong robustness on novel classes.
In contrast, our proposed Robust-ProL achieves satisfying robustness on both base classes and novel ones, with over 20\% accuracy increase on novel classes on various OOD datasets. Besides, there is an interesting observation that IcoL-Random baseline performs better on novel classes than on base classes, which indicates that the demonstration from disjoint classes is better than random examples from base classes. We leave the further exploration of this observation in future work.

\noindent\textbf{Comparison to Robustness of Unsupervised Performance on VLMs.} In Sec.~\ref{sec:basec}, we compare the robustness of IcoL and ProL with the robustness of unsupervised learning on base classes. Since standard supervised learning could hardly generalize to novel classes, we compare the model performance on novel classes with unsupervised (i.e., zero-shot) performance of VLM. As shown in Tab.~\ref{tab:novel}, The unsupervised baseline performs similarly on base classes and novel classes since no supervision is provided. Compared with the unsupervised baseline, Robust-ProL performs much better on base classes and achieves comparable robustness on novel classes on most OOD datasets.

\begin{table}[t]
\begin{center}
\footnotesize
\setlength\tabcolsep{0.18cm}
\begin{tabular}{l| cc | cc | cc }
\toprule
\multirow{2}{*}{Datasets} & \multicolumn{2}{c|}{Man-ProL}  &  \multicolumn{2}{c|}{Co-ProL} & \multicolumn{2}{c}{Robust-ProL} \\
\cline{2-7}
 & 8 Cls  &  16 Cls  & 8 Cls  &  16 Cls & 8 Cls  &  16 Cls \\
\midrule
ImageNet    & 58.25  & 32.13  & 59.75 & 46.13 &   73.25  &  58.5\\
 \midrule
ImageNet-V2  & 56.67  & 32.92 & 65.41 & 46.04 &   71.25 &  57.5 \\
\midrule
ImageNet-R  & 57.14  & 26.94  & 41.12 & 31.18 &   64.27  & 44.90 \\
\midrule
ImageNet-C & 42.33  & 19.73   & 39.23  & 26.14 &  60.28  & 39.42 \\
\midrule
ImageNet-S  & 56.54  & 32.39  & 47.65 & 37.81 &   64.94  & 51.35 \\
\midrule
ImageNet-A & 38.06 & 18.73  & 32.47 & 20.58 &  53.76 & 31.79 \\
\bottomrule
\end{tabular}
\end{center}
\caption{We compare Robust-ProL with Man-G-Context and Co-ProL on novel classes of all datasets, which are inspired by CoOp and CoCoOp, respectively. While Co-ProL improves over Man-ProL on in-distribution datasets, it underperforms Man-G-Prompt on OOD datasets. Robust-ProL clearly outperforms both Man-ProL and Co-ProL.} 
\label{tab:sota_prompt}
\end{table}

\subsection{Robust-ProL vs SOTA ProL}
ProL has also been intensively studied in natural language understanding and in contrastive learning-based VLMs.
On contrastive VLM, part of the text (called context) is optimized with a learning-based method, dubbed Context Optimization (CoOp)~\cite{zhou2022learning}. To improve the performance on open-set classes, they propose to condition the CoOp on visual features, which is dubbed CoCoOp~\cite{zhou2022conditional}. Besides, the popular Adapter method~\cite{zhang2021tip,gao2021clip} has also been explored by adding adaptation layers in VE.

Considering that only a few examples are available in our in-context setting, we mainly compare Robust-ProL with Man-ProL (inspired by CoOp) and Co-ProL (inspired by CoCoOp) since Adapter methods can easily overfit to the small data. The three methods perform similarly on base classes across all datasets (See Appendix C) but our Robust-ProL significantly outperforms both context learning variants on novel classes, as shown in Tab.~\ref{tab:sota_prompt}. The class-conditioning of context learning in Co-ProL improves the performance over Man-ProL on novel classes of in-distribution datasets, which is consistent with findings of previous work~\cite{zhou2022conditional}. However, Co-ProL underperforms Man-ProL on OOD datasets implying the lack of robustness of class-conditional prompt learning.

\begin{table}[t]
\begin{center}
\footnotesize
\setlength\tabcolsep{0.14cm}{
\begin{tabular}{l| c| cccc}
\toprule
 & ProL & R-ProL & R-ProL-SS(384) & R-ProL-SS(299)  \\
\midrule
Computation & $\times 1$ & $\times 1$\ & $\times 1$\ & $\times 1$  \\
\midrule
\midrule
ImageNet & 95.75 &97.75 &  97.22 $\;( \downarrow  0.53)$ & 95.5 $\;( \downarrow  2.25)$ \\
 \midrule
ImageNet-V2 &97.08 & 98.75 & 95.02 $\;( \downarrow  3.73)$ & 92.5 $\;( \downarrow  6.25)$  \\
\midrule
ImageNet-R & 64.26 & 74.55 & 70.59 $\;( \downarrow  3.96)$ & 63.86 $\;( \downarrow  13.69)$ \\
\midrule
ImageNet-C & 66.59 & 72.61 & 70.26 $\;( \downarrow  2.35)$ & 64.15 $\;( \downarrow  8.46)$ \\
\midrule
ImageNet-S & 71.32 & 84.56  & 83.31 $\;( \downarrow  1.25)$ & 75.25 $\;( \downarrow  9.33)$ \\
\midrule
ImageNet-A & 65.84 & 68.81 & 67.82 $\;( \downarrow  0.99)$ &  54.21 $\;( \downarrow 14.60 )$ \\
\bottomrule
\end{tabular}}
\end{center}
\caption{Multi-scale features in Robust-ProL: the robustness on base classes of R-ProL-SS with a repetition of a single scale drops to different degrees compared to Robust-ProL. 
The repetition of visual features also brings a gain when comparing R-ProL-SS to ProL. Robustness on novel classes can be found in Appendix D.}
\label{tab:single_scale} 
\end{table}

\subsection{Further Understanding of Robust-ProL}
In this section, we further perform ablations to better understand the robustness benefits of Robust-ProL by comparing it with 1) ProL with a single-scaled test image and 2) ensembling methods of ProL.

\noindent\textbf{Multi-scale Features of Robust Prompt}.
To test the effectiveness of multi-scale features, we apply a single scale of image three times in Robust-ProL. That is, we keep the visual feature tokens in Robust-ProL at the same scale and denote it as R-ProL-SS(scale). For example, R-ProL-SS(384) means a scale of 384 is used.

As shown in Tab.~\ref{tab:single_scale}, we can see that the accuracy of R-ProL-SS decreases to different degrees compared to Robust-ProL. This indicates that multi-scale features integrated by Robust-ProL indeed contribute to the robustness improvement. Besides, note that all the methods in Tab.~\ref{tab:single_scale} take a similar computational cost since the language model dominates the cost and the input tokens of the language model are appended to a fixed length.

In addition, the superior performance of R-ProL-SS(384) over ProL suggests that a simple repetition of input image can also result in robustness benefit. To further understand the observation, in the next experiment, we compare Robust-ProL with traditional ensembling~\cite{dietterich2000ensemble} of ProL with a single test image.

\begin{table}[t]
\begin{center}
\footnotesize
\setlength\tabcolsep{0.08cm}{
\begin{tabular}{l|ccc|c}
\toprule
 & ProL & Ensemble-SS & Ensemble-MS & Robust-ProL  \\
\midrule
Computation & $\times 1$ & $\times 3$ & $\times 3$ & $\times 1$  \\
\midrule
\midrule
ImageNet & 95.75 & 98.5  $\;( \tiny \uparrow 2.75 )$ & 98 $\;( \tiny \uparrow 2.25)$ & 97.75  \\
 \midrule
ImageNet-V2 & 97.08 & 97.5 $\;( \tiny \uparrow 0.42)$ & 97.5 $\;( \tiny \uparrow 0.42)$ & 98.75   \\
\midrule
ImageNet-R & 64.26 & 75.14 $\;( \tiny \uparrow 10.88 )$ & 73.17 $\;( \tiny \uparrow 8.91 )$ & 74.55  \\
\midrule
ImageNet-C & 66.59 & 69.28 $\;( \tiny \uparrow 2.69)$ & 70.03 $\;( \tiny \uparrow 3.44)$ & 72.61  \\
\midrule
ImageNet-S & 71.32 & 84.55 $\;( \tiny \uparrow 13.23)$ & 82.59 $\;( \tiny \uparrow 11.27)$ & 84.56 \\
\midrule
ImageNet-A & 65.84 & 60.91 $\;(\downarrow 4.93)$ & 62.87 $\;(\downarrow 3.03 )$ & 68.81 \\
\bottomrule
\end{tabular}}
\end{center}
\caption{Ensemble of ProL improves the model performance on base classes of most datasets. Our R-prompt outperforms the two ensembling methods with the same computational cost as ProL, while the ensembling methods require N=3 times more. More scores on novel classes are in Appendix E.} 
\label{tab:ensemble}
\end{table}

\noindent\textbf{Ensembling Effect of Robust Prompt Learning}.  
We compare our Robust-ProL with two ensembling methods of ProL: 1) Ensemble-SS: We simply run ProL learning three times with different seeds and make a prediction based on the average of model losses. A single image of the same scale (i.e., 384) is applied to each ProL. 2) Ensemble-MS: An ensemble of three ProL, each of which is equipped with a different image scale. The three scales used in Ensemble-MS are the same as in Robust-ProL.

In Tab.~\ref{tab:ensemble}, we can observe that both ensembling methods of ProL outperform a single ProL, while our Robust-ProL achieves stronger robustness than traditional ensembling of ProL.  Note that our Robust-ProL can also be considered as an efficient form of ensembling since it requires the same computational cost as ProL, while the traditional ensembling methods requires N(=3) times more.

\section{Concluding Remarks}
In this work, we studied the robustness of in-context learning 
(IcoL) and prompt learning (ProL) on Vision-Language Models (VLMs). 
We showed that prompt learning methods generally improve model robustness on base classes (seen in the exemplar). However, ProL does not perform robustly on novel unseen classes. 

We proposed Robust-ProL that combines visual features at multiple different scales in the prompt.
We showed that the proposed Robust-ProL achieves higher robustness on both base classes and novel ones, outperforming all existing prompting methods. This is likely because multiple-scale images can help to capture the different features of an image, which can make it more difficult for overfitting spurious correlations.


\section{Limitations}
One of the shortcomings of this study is that we only performed experiments using MEGMA~\cite{eichenberg2021magma}, as the only publicly accessible VLM with frozen LLM and VE, as of the writing of this paper. It is left open as to how much of the conclusions of this paper generalize across various VLMs, and specifically whether the proposed robust prompt learning that integrates multi-scale images generalizes to provide a significant robustness boost on other VLMs as well.

Secondly, we chose the classes used for the support set of the prompt learning, and the images used in the exemplar from each of the chosen classes uniformly at random and fixed them throughout the study. We also performed learning for each method only once on those selected classes. As such, it is expected that the reported numbers could vary based on the choice of the classes and the particular images within each class.
 Moreover, we chose the novel (unseen) classes for testing generalization uniformly at random from the unseen classes, and fixed them throughout the study. As such the reported accuracies on novel classes are subject to variations based on how such classes might be related to the base classes. A more rigorous study of these issues is left open for future work.
{\small
\bibliographystyle{ieee_fullname}
\bibliography{arxiv}
}

\newpage
\appendix
\section{Appendix: Names of Base and Novel Classes}
We first select 32 classes randomly from the classes shared by the six datasets. As base classes, the selected classes are used for the classification task of 32 classes. For the classification of 8, 16 classes, we use the first 8 classes, the first 16 classes from the 32 selected classes, respectively. Concretely, the selected classes are as follows: 

8-class classification: \textit{'ambulance', 'bell pepper', 'fox squirrel', 'basketball', 'grasshopper', 'tarantula', 'centipede', 'accordion'}.

16-class classification: \textit{'ambulance', 'bell pepper', 'fox squirrel', 'basketball', 'grasshopper', 'tarantula', 'centipede', 'accordion', 'sax', 'porcupine', 'hotdog', 'rugby ball', 'parachute', 'baboon', 'vulture', 'bow tie'}.

32-class classification: \textit{'ambulance', 'bell pepper', 'fox squirrel', 'basketball', 'grasshopper', 'tarantula', 'centipede', 'accordion', 'sax', 'porcupine', 'hotdog', 'rugby ball', 'parachute', 'baboon', 'vulture', 'bow tie', 'Rottweiler', 'mantis', 'lion', 'cucumber', 'broccoli', 'flamingo', 'jellyfish', 'ant', 'broom', 'bison', 'mushroom', 'American egret', 'school bus', 'African chameleon', 'ladybug', 'volcano'}.

The following 8 classes are taken as novel classes. Please note that the novel classes are randomly selected from the classes shared by the six datasets and excluded from the 32 base classes.

8-class classification: \textit{'cheeseburger', 'candle', 'monarch', 'goldfinch', 'hermit crab', 'banana', 'drake', 'canoe'}. 

\section{Appendix: Ablation on Hyperparameters of Learnable Context}
Learning context is based on the examples from a supporting set assumed by in-context learning. Since learning on a few examples can leads to overfitting, we conduct experiments to study the impact of the following two factors: the number of learnable tokens and training epochs. 

\begin{figure*}[h]
    \centering
    \begin{subfigure}[b]{0.22\textwidth}
        \includegraphics[scale=0.4]{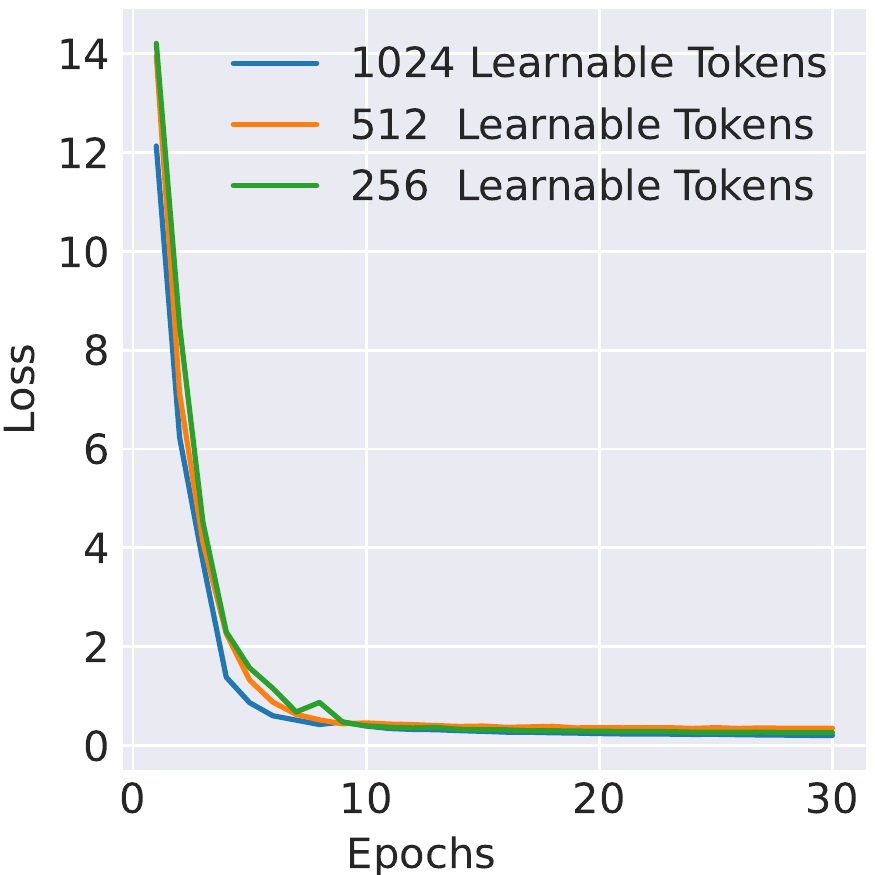}
    \caption{L-Context Learning}
    \end{subfigure} \hspace{2mm}
    \begin{subfigure}[b]{0.22\textwidth}
        \includegraphics[scale=0.4]{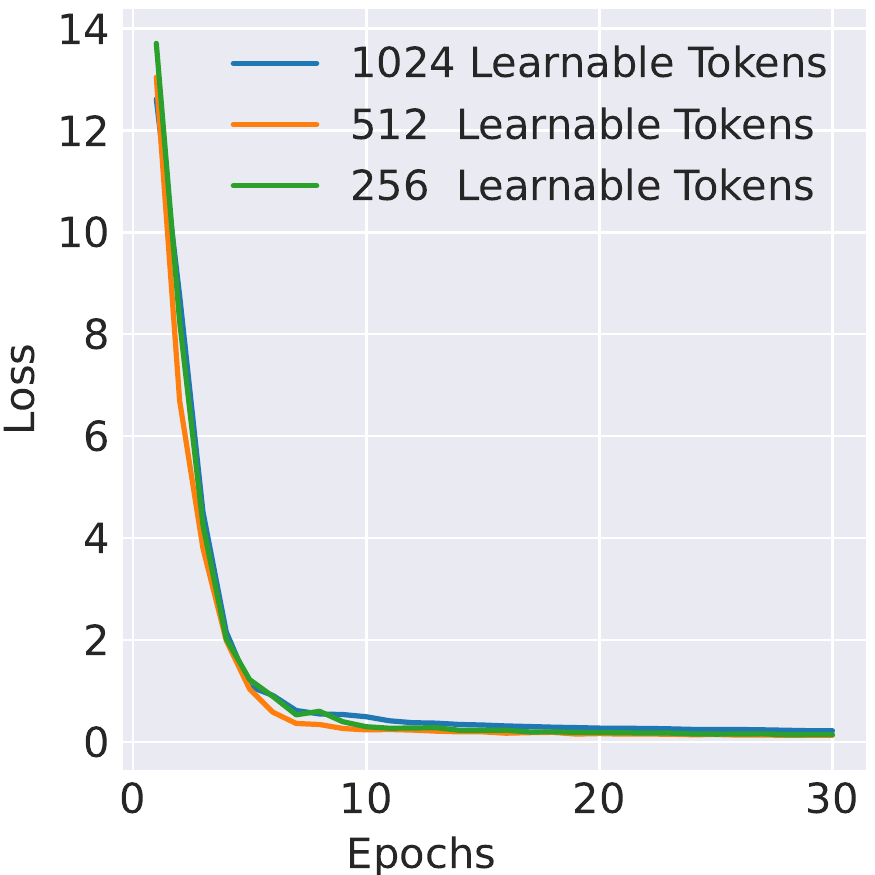}
        \caption{L-Context Test}
    \end{subfigure} \hspace{2mm}
    \begin{subfigure}[b]{0.22\textwidth}
        \includegraphics[scale=0.4]{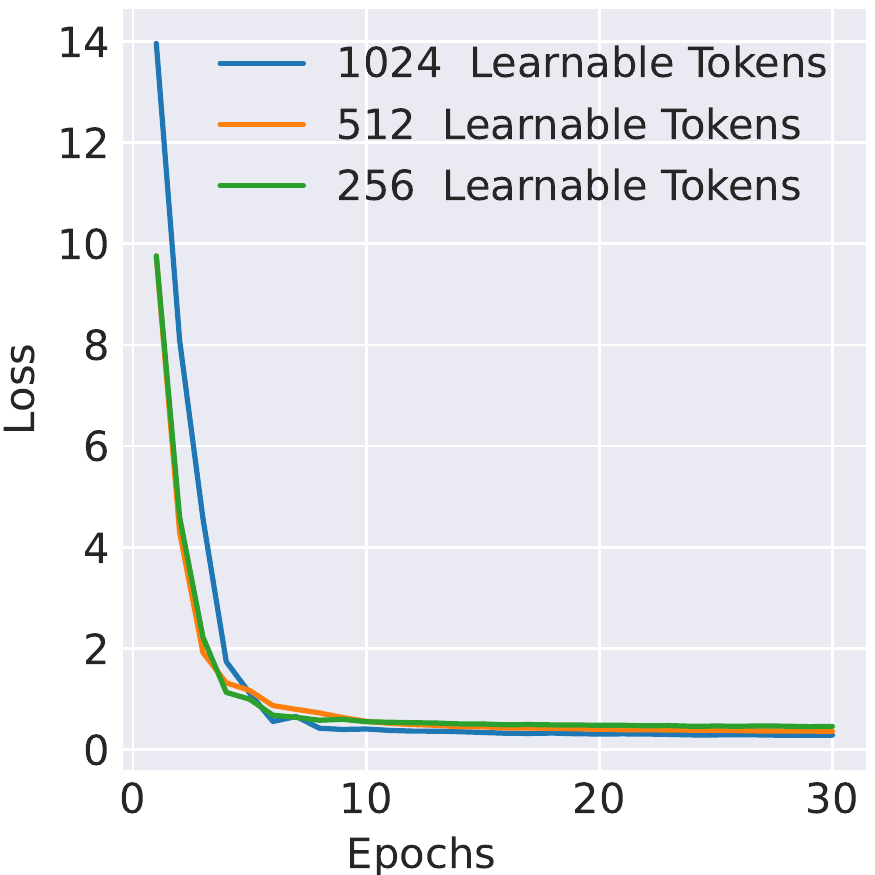}
        \caption{R-Context Learning}
    \end{subfigure} \hspace{2mm}
    \begin{subfigure}[b]{0.22\textwidth}
        \includegraphics[scale=0.4]{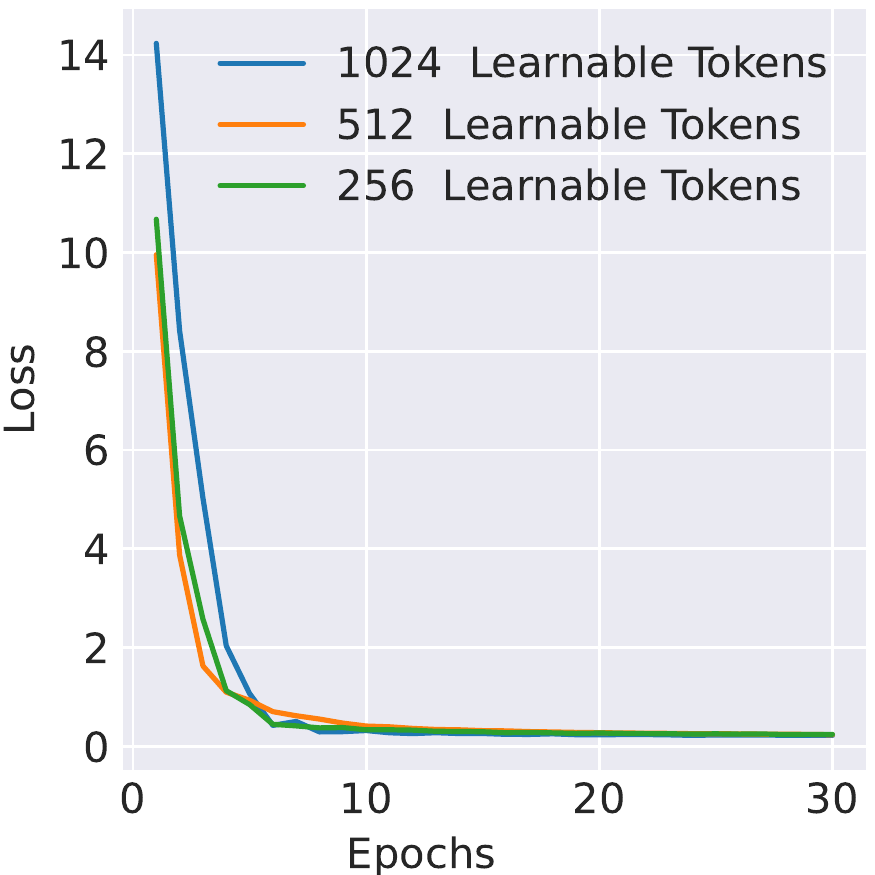}
        \caption{R-Context Test}
    \end{subfigure} 
    \caption{We visualize both training and test loss during prompt learning. Different numbers of learnable tokens are also considered. We can observe that no overfitting exists during learning, and both L-Context and R-Context learning are not sensitive to the number of learnable tokens.}
    \label{fig:ab_tokens_epochs}
\end{figure*}

\vspace{0.05cm}
We train both L-Context and R-Context with the different numbers of learnable tokens and visualize the training and test losses in Fig. \ref{fig:ab_tokens_epochs}. As shown in the figure, no overfitting is observed in both cases, even when we train them 100 epochs. It is not surprising since both VE and LM of VLM are frozen in the prompt learning process. Besides, we can also observe that the learning process is not sensitive to the number of learnable tokens. The training and test losses can be rapidly reduced to nearly zero in both cases.

\newpage

\vspace{0.1cm}
\section{Appendix: Performance of Man-L-Context, Co-L-Context and R-Context on Base Classes}
We compare R-Context with Man-L-Context (inspired by CoOp) and Co-L-Context (inspired by CoCoOp). The three methods perform similarly on base classes across all datasets.

\begin{table*}[h]
\begin{center}
\footnotesize
\setlength\tabcolsep{0.06cm}
\begin{tabular}{c| ccc| ccc| ccc| ccc| ccc| ccc}
\toprule
 Datasets & \multicolumn{3}{c|}{ImageNet} & \multicolumn{3}{c|}{ImageNet-V2} & \multicolumn{3}{c|}{ImageNet-R} &  \multicolumn{3}{c|}{ImageNet-C} &  \multicolumn{3}{c|}{ImageNet-S} & \multicolumn{3}{c}{ImageNet-A} \\
\midrule
No. of Classes & 8 & 16 & 32  &  8 & 16 & 32  & 8 & 16 & 32 & 8 & 16 & 32 & 8 & 16 & 32 & 8 & 16 & 32 \\
 \midrule
CoOp  & 98.15 & 94.60 &  91.00   & 98.75 & 94.38 & \textbf{93.02}  & 76.34 & 66.26 & 63.23 & \textbf{75.14} & 57.74 &  57.26  & \textbf{85.54} & 81.34 & \textbf{76.95}  & 67.57 & 49.13 & \textbf{46.95} \\
\midrule
CoCoOp  & \textbf{98.25} & 94.63 &  90.19   & 97.92 & 94.17 & 91.56  & \textbf{76.74} & 67.04 &  65.08  & 67.36 & 59.05 &  54.77  & 75.00 & 76.69 & 76.15  & 67.57 & 49.13 & 38.91 \\
\midrule
R-Context  &  97.75 & \textbf{95.88} & \textbf{91.69} & \textbf{98.75}  & \textbf{95.63} & 91.88 & 74.55 & \textbf{74.36} &  \textbf{65.33} & 72.61 & \textbf{68.58} & \textbf{58.42}  & 84.56 & \textbf{83.68} & 75.23 & \textbf{68.81} & \textbf{59.78} & 42.6\\
\bottomrule
\end{tabular}
\end{center}
\caption{Performance of Man-L-Context, Co-L-Context and R-Context on Base Classes on all Six Datasets.}
\label{tab:sota_models}
\end{table*}

\vspace{0.4cm}
\section{Appendix: Ablation on Multi-scale Feature of Prompt Learning}
The accuracy of R-Context-SS decreases to different degrees compared to R-Context. This indicates that multi-scale features indeed contribute to robustness improvement. Note that all the methods take the same cost.

\begin{table*}[h]
\begin{center}
\footnotesize
\setlength\tabcolsep{0.06cm}
\begin{tabular}{c| ccc| ccc| ccc| ccc| ccc| ccc}
\toprule
 Datasets & \multicolumn{3}{c|}{ImageNet} & \multicolumn{3}{c|}{ImageNet-V2} & \multicolumn{3}{c|}{ImageNet-R} &  \multicolumn{3}{c|}{ImageNet-C} &  \multicolumn{3}{c|}{ImageNet-S} & \multicolumn{3}{c}{ImageNet-A} \\
\midrule
No. of Classes & 8 & 16 & 32  &  8 & 16 & 32  & 8 & 16 & 32 & 8 & 16 & 32 & 8 & 16 & 32 & 8 & 16 & 32 \\
 \midrule
R-Context-SS  & 97.22 & \textbf{95.92} &  \textbf{95.25}   & 95.02 & 94.97 & 90.25  & 70.59 & 71.82 &  \textbf{70.06}  & 70.26 & 68.27 & \textbf{66.34} & 83.16 & 77.30 & \textbf{78.14}  & 67.82 & \textbf{60.73} &  \textbf{65.84} \\
\midrule
R-Context  &  \textbf{97.75} & 95.88 & 91.69 & \textbf{98.75}  & \textbf{95.63} & \textbf{91.88} & \textbf{74.55} & \textbf{74.36} &  65.33 & \textbf{72.61} & \textbf{68.58} & 58.42 & \textbf{84.56} & \textbf{83.68} & 75.23 & \textbf{68.81} & 59.78 & 42.6\\
\bottomrule
\end{tabular}
\end{center}
\caption{Multi-scale features in R-Context: the robustness on base classes of R-Context-SS with a repetition of a single scale drops to different degrees compared to R-Context in most cases.}
\label{tab:sota_models}
\end{table*}

\vspace{0.4cm}
\section{Appendix: Ensemble Performance of L-Context on Novel Classes across All Datasets}
We claim that both ensembling methods of L-Context outperform a single L-Context, while our R-Context achieves stronger robustness than traditional ensembling. Our experiments on novel classes also support our claim, as shown in Tab. \ref{tab:ensemble}.

\begin{table}[h]
\begin{center}
\footnotesize
\setlength\tabcolsep{0.5cm}
\begin{tabular}{l|ccc|c}
\toprule
 & L-Context & Ensemble-SS & Ensemble-MS & R-Context  \\
\midrule
Computation & $\times$ 1 & $\times$ 3 & $\times$ 3 & $\times$ 1  \\
\midrule
\midrule
ImageNet & 43.5 & 56.25 $\;( \uparrow 12.75)$ & 60.75 $\;( \uparrow 17.25)$ & \textbf{73.25} \\
 \midrule
ImageNet-V2 & 45.42 & 64.17 $\;( \uparrow  18.75)$ & 57.92 $\;( \uparrow 12.5)$ & \textbf{71.25} \\
\midrule
ImageNet-R & 39.56 & 42.87 $\;( \uparrow 3.31)$ & 47.21 $\;( \uparrow 7.65)$ & \textbf{64.27} \\
\midrule
ImageNet-C & 39.23 & 43.01 $\;( \uparrow 3.78)$ & 45.32 $\;( \uparrow 6.09)$ & \textbf{60.28} \\
\midrule
ImageNet-S & 40.99 & 40.02 $\;( \downarrow 0.97)$ & 46.91 $\;( \uparrow 5.92)$ & \textbf{64.94} \\
\midrule
ImageNet-A & 20.56 & 19.78 $\;( \downarrow 0.78)$ & 25.81 $\;( \uparrow 5.25)$ & \textbf{53.76} \\
\bottomrule
\end{tabular}
\end{center}
\caption{Ensemble Performance of L-Context on novel classes across all six datasets. Ensembling methods of L-Context outperform a single L-Context, but underperform R-Context.}
\label{tab:ensemble}
\end{table}

\end{document}